\documentclass{article}
\usepackage{spconf}
\usepackage{amsmath,amssymb,amsfonts}
\usepackage{algorithmic}
\usepackage{graphicx}
\usepackage[ruled]{algorithm2e}
\usepackage{booktabs}
\usepackage{textcomp}
\usepackage{hyperref}

\usepackage{enumitem}
\setlist{nosep, leftmargin=14pt}

\usepackage{mwe} 

\title{Text2Cohort: Facilitating Intuitive Access to Biomedical Data with Natural Language Cohort Discovery}
\name{Pranav Kulkarni, Adway Kanhere, Paul H. Yi, Vishwa S. Parekh\thanks{Corresponding author: vparekh@som.umaryland.edu}}
\address{University of Maryland Medical Intelligent Imaging (UM2ii) Center,\\University of Maryland School of Medicine, Baltimore, MD 21201}
\begin{document}
%
\maketitle
\begin{abstract}
The Imaging Data Commons (IDC) is a cloud-based database that provides researchers with open access to cancer imaging data, with the goal of facilitating collaboration. However, cohort discovery within the IDC database has a significant technical learning curve. Recently, large language models (LLM) have demonstrated exceptional utility for natural language processing tasks. We developed Text2Cohort, a LLM-powered toolkit to facilitate user-friendly natural language cohort discovery in the IDC. Our method translates user input into IDC queries using grounding techniques and returns the query's response. We evaluate Text2Cohort on 50 natural language inputs, from information extraction to cohort discovery. Our toolkit successfully generated responses with an 88\% accuracy and 0.94 F1 score. We demonstrate that Text2Cohort can enable researchers to discover and curate cohorts on IDC with high levels of accuracy using natural language in a more intuitive and user-friendly way.
\end{abstract}
\section{Introduction}
\label{sec:intro}

The National Cancer Institute's Imaging Data Commons (IDC) is a cloud-based data commons that provides researchers with open access to large-scale cancer imaging datasets and tools for analysis, with the goal of facilitating the sharing of imaging data and promoting collaboration in the field of medical imaging research \cite{fedorov2021nci,grossman2016case,jiang2022big}. The IDC is hosted on the Google Cloud Platform (GCP) with the DICOM metadata across all IDC collections indexed in the form of a BigQuery relational database to enable powerful queries and cohort discovery for any IDC user. However, curating cohorts by querying the BigQuery database requires extensive knowledge of the data schema, knowledge of Structured Query Language (SQL), and a sandbox environment with Python to access, curate, and download the imaging data on IDC. This is a major bottleneck for users without extensive knowledge of the data schema or technical skills to effectively query these datasets and curate multi-collection cohorts. 

\begin{figure}[!t]
    \centering
    \includegraphics[width=\linewidth]{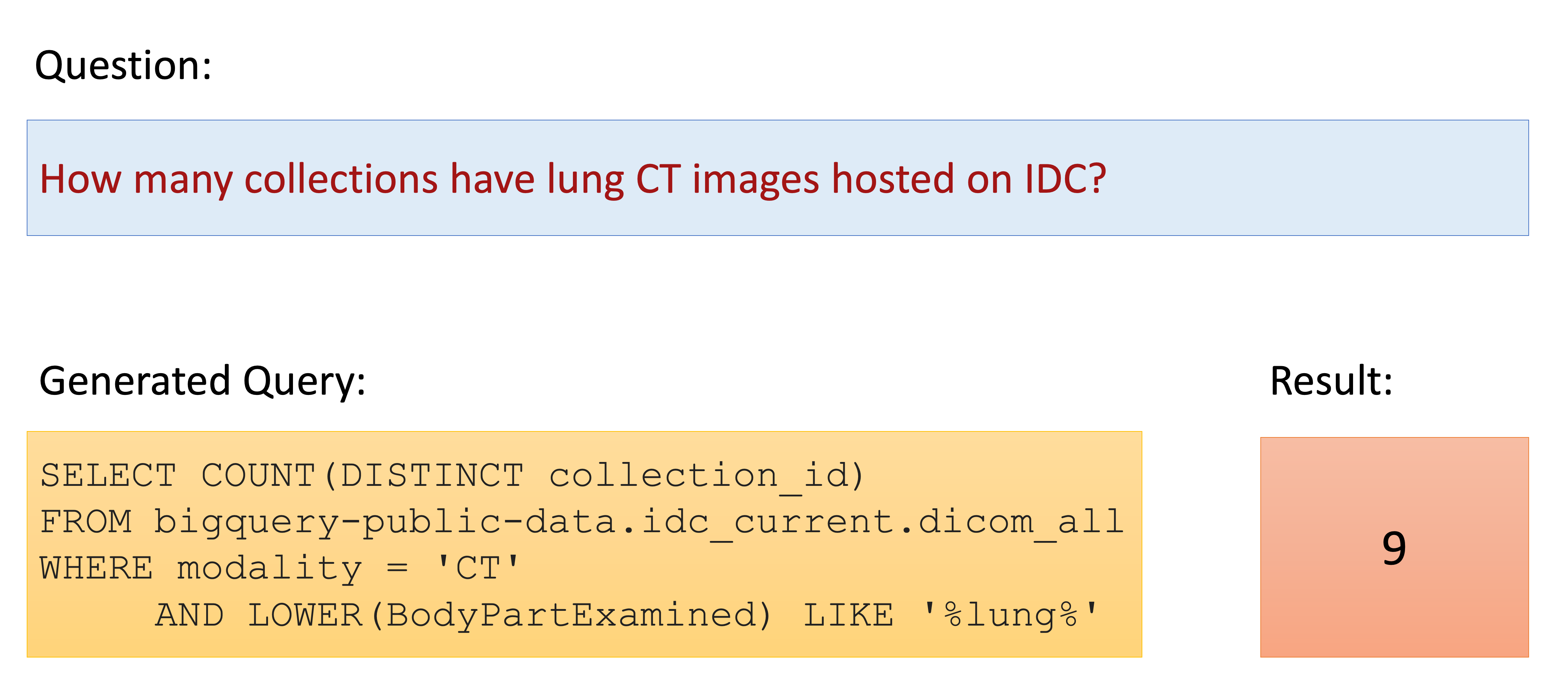}
    \caption{The Text2Cohort toolkit on an example natural language user input. Text2Cohort first transforms the user input into a query, uses the generated query to query the BigQuery database, and returns the response back to the user.}
    \label{fig:figure0}
\end{figure}

Recently, large language models (LLMs) with billions of parameters pre-trained on enormous corpus' of text data, such as news articles, books, and web pages on the entire internet are revolutionizing the field of natural language processing with exceptional ability to understand and respond to natural language queries \cite{khan2022transformers,brown2020language,liu2023pre}. At their core, LLMs learn to identify patterns and relationships between words and phrases in the text and develop an understanding of the structure and grammar of language. LLM models like GPT-3.5 have previously demonstrated exceptional capabilities in generating SQL queries from natural language input in zero- and few-shot scenarios \cite{sun2023battle,rajkumar2022evaluating,kumar2022deep,yang2022empirical,dong2023c3}. However, LLMs occasionally hallucinate generating incorrect queries with syntactical or semantic errors, requiring expert understanding to identify and correct these errors \cite{sun2023battle}. These errors could significantly limit the translational capability of LLMs for use by non-experts. Furthermore, executing the generated SQL queries would still need a sandbox environment with Python to access the data and retrieve query's response. 

To that end, we developed Text2Cohort, an end-to-end natural language cohort-discovery toolkit that would enable researchers to interact with and discover cohorts from multiple collections on IDC simultaneously in a more intuitive and user-friendly way, thus eliminating the learning curve associated with writing SQL queries or understanding the IDC database schema (Fig. \ref{fig:figure0}). Text2Cohort uses self-supervised autocorrection to iteratively identify and correct hallucinations in the generated SQL queries and return the query's correct response to the user. We evaluated our framework on 50 natural language queries ranging from information extraction to cohort curation with varying levels of complexity to evaluate for hallucination correction and response generation.

\section{Materials and Methods} \label{sec:methods}
\subsection{Text2Cohort}

\begin{figure}[!t]
    \centering
    \includegraphics[width=\linewidth]{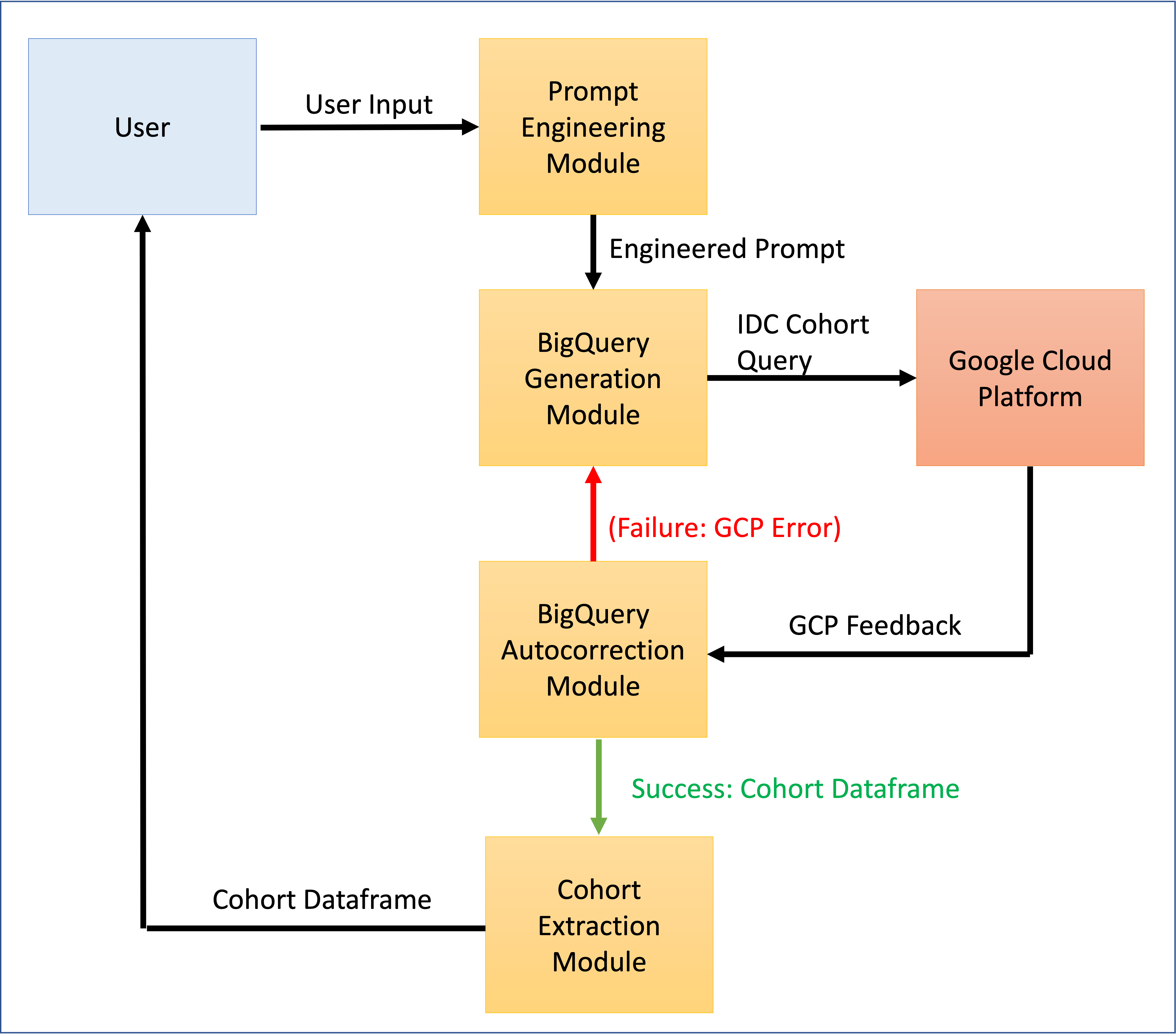}
    \caption{Illustration of the Text2Cohort toolkit.}
    \label{fig:figure1}
\end{figure}

The Text2Cohort toolkit is built using GPT-3.5 and consists of four major components: (1) prompt engineering, (2) BigQuery generation, (3) BigQuery autocorrection, and (4) cohort extraction, as illustrated in Fig. \ref{fig:figure1}.

\subsubsection{Prompt Engineering} 

While GPT-3.5 provides a sophisticated interface for natural language processing, it is crucial ground the model via prompt engineering to provide contextual information, focus the model's responses for the task at hand, and minimize hallucinations. In other words, this enables a zero-shot fine-tuning of the model's capabilities for the given task. In Text2Cohort, we utilize prompt engineering to prime GPT-3.5 for query generation as follows:
    
\begin{enumerate}
    \item Query from the public BigQuery database, which contains DICOM metadata for all collections hosted by the IDC.
    \item Queries should be as specific as possible without providing explanations behind responses to reduce time taken to generate queries.
    \item Queries must be generated enclosed within fixed delimiters to simplify query extraction.
    \item Queries should utilize regular expressions in queries to prevent exact matches, thus resulting in a more generalizable query structure.
\end{enumerate}

\subsubsection{BigQuery Generation} 

Once GPT-3.5 is primed for query generation, the user can enter a free-text query such as "How many male brain MRI images are hosted on IDC?" or "I want all images in the NSCLC Radiomics collection". This input is fed to the model for interpretation and query generation. The resultant query is extracted from the response and queried to IDC's BigQuery database using the GCloud SDK's BigQuery client.

\begin{figure}[!b]
    \centering
    \includegraphics[width=\linewidth]{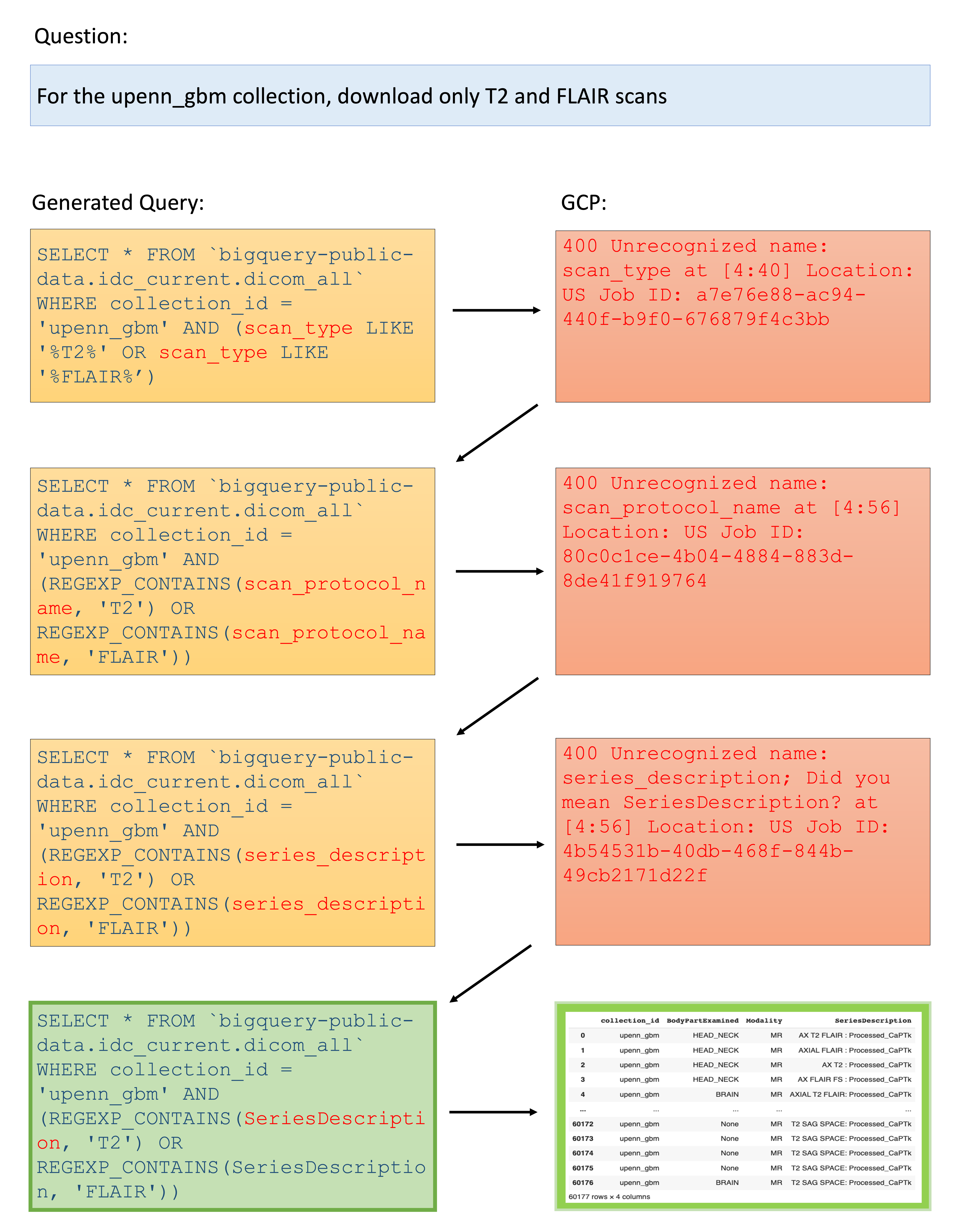}
    \caption{Illustration of the autocorrection pipeline for an example user input. The example demonstrates how the autocorrection pipeline recursively autoengineers the prompt to guide the LLM towards using the keyword SeriesDescription to filter different MRI sequences.}
    \label{fig:figure2}
\end{figure}

\begin{table*}[!ht]
    \caption{An example of Text2Cohort's generated queries and responses on five natural language user inputs.}
    \centering
    \includegraphics[width=0.9\linewidth]{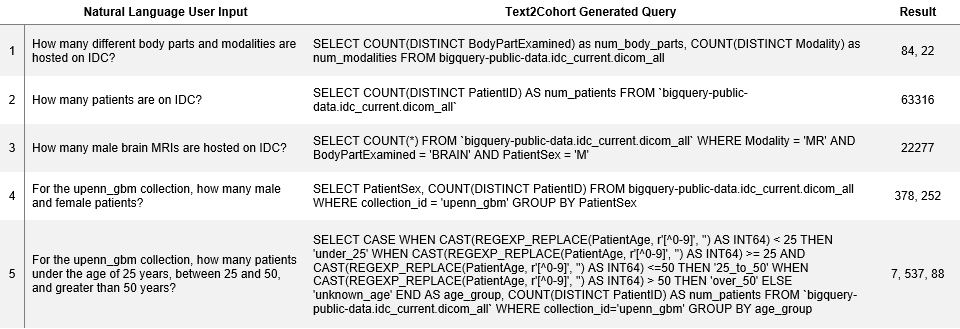}
    \label{fig:results1}
\end{table*}

\subsubsection{BigQuery Autocorrection} 

In many cases, GPT-3.5 generates an incorrect query containing errors that can be classified under two groups: (1) syntax errors and (2) semantic errors. Syntax errors can occur due a typo or an incorrectly labeled field, while semantic errors can occur due to an incorrect interpretation of the input text. Text2Cohort implements an autocorrect pipeline to address both errors. As shown in Fig. \ref{fig:figure2}, the autocorrection pipeline ingests the associated error message when the generated query is incorrect and passes back to GPT-3.5 to interpret the error and attempt to fix it. Text2Cohort's autocorrect pipeline is implemented recursively to attempt query autocorrection until the query is executed successfully. However, there are a few limitations to our autocorrection pipeline: (1) In some cases, semantic errors may not be corrected if the underlying query is executed successfully, thus resulting in an incorrect response, and (2) we limit autocorrection to at most 10 attempts to prevent token usage from exceeding OpenAI's API token limit.

\subsubsection{Cohort Extraction} 

The cohort extraction component of the Text2Cohort toolkit uses the generated and autocorrected query to query the BigQuery database and extract the resultant table as a Pandas dataframe in Python.

\subsection{Experiments}

To initiate our study, we curated a dataset of $n=50$ natural language user inputs to evaluate Text2Cohort, with queries ranging from information extraction to cohort discovery, like "How many male and female patients are present in the NSCLC Radiomics dataset?" and "Please curate a dataset of all male brain MRI patients hosted on IDC". The toolkit was evaluated on these natural language user inputs and the resultant queries and responses were expert-verified by consensus between two computer scientists as either correct or incorrect; disagreements were adjudicated by a third computer scientist. The efficacy of the natural language toolkit was consequently measured by its accuracy and F1 scores across all user inputs. For user inputs that generated incorrect queries and responses, the query was corrected by an expert and the Levenshtein distance between the corrected query and the incorrect query was calculated. In short, the Levenshtein distance measures the minimum number of character modifications to change one string into another 

\section{Results} \label{sec:results}

\begin{table*}[!t]
    \centering    
    \caption{Natural language user inputs to which Text2Cohort generated incorrect responses. Expert corrected queries are provided. Errors highlighted in red and corrections highlighted in blue.}
    \includegraphics[width=0.9\linewidth]{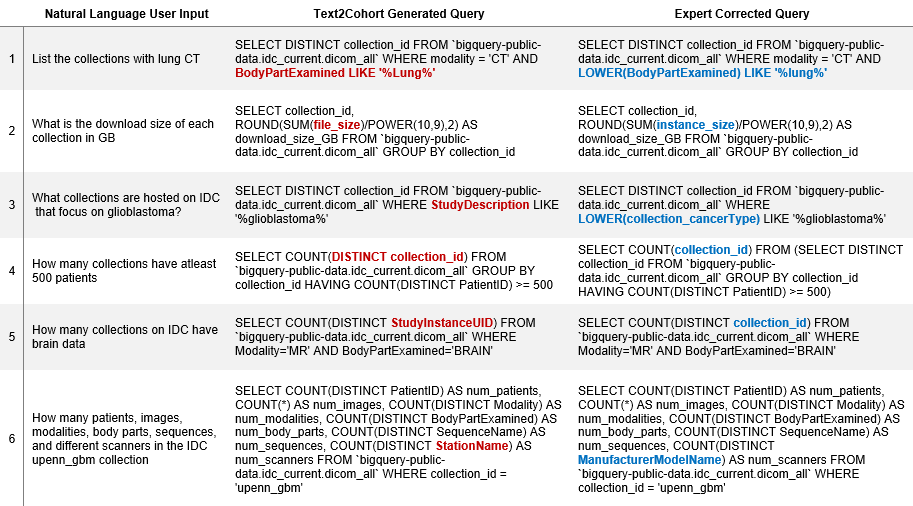}
    \label{fig:results3}
\end{table*}

Our results indicate that on all 50 curated natural language user inputs, across information extraction and cohort discovery tasks, Text2Cohort demonstrates excellent performance with an accuracy of 88\% and F1 score of 0.94 in generating correct responses to the user inputs. The performance of the toolkit on an example set of information extraction and cohort discovery queries is illustrated in Table \ref{fig:results1}. In other words, the toolkit generated correct queries and responses to 44 out of 50 user inputs (88\%) but failed to do so for six user inputs (12\%), as shown in Table \ref{fig:results3}. Out of these six incorrect responses, one (17\%) resulted in a query that exceeded the maximum number of autocorrection attempts, while five (83\%) failed due to semantic errors within the generated queries. Furthermore, out of the five responses containing semantic errors, three (60\%) failed due to the generated query using an incorrect field for the task. These six incorrect queries were manually corrected by an expert with $12.83 \pm 5.81$ character-edits determined by the Levenshtein distance between the corrected and incorrect queries (Table \ref{fig:results3}). In short, our results demonstrate that despite failing to correct 10\% of all queries due to semantic errors, Text2Cohort has the ability to generate queries with correct structure and autocorrect syntax errors within them with a 98\% success rate. 

\section{Discussion} \label{sec:discussion}

Text2Cohort yields excellent results in translating natural language user input into powerful database queries, and subsequently into responses, through prompt engineering and autocorrection. Furthermore, it demonstrates the utility of LLMs to facilitate natural language information extraction and cohort discovery by enabling a more intiutive and user-friendly interface to the IDC and other similar databases. Furthermore, it eliminates the need for a technical understanding of databases and the underlying data schema. In other words, our work demonstrates that not only does Text2Cohort revolutionize how researchers can discover cohorts, interact with, and access imaging data hosted on the IDC, it also democratizes access to the IDC.

However, the utility of the Text2Cohort toolkit is limited due to a few bottlenecks. Firstly, Text2Cohort requires an understanding of the entire data schema to reach its full potential. In our study, we observed that all incorrect responses were due to the lack of an understanding of the data schema (e.g., incorrectly interpreting collections as studies). While it is evident that LLM can encode facts, they are prone to fabricating facts without appropriate supervision \cite{choi2023chatgpt}. Text2Cohort's autocorrection pipeline functions as weak supervision by allowing the model to interpret and correct any errors generated while querying. However, autocorrection is limited in its utility when handling semantic errors. For example, a query containing semantic errors may successfully execute, bypass autocorrection, and return an incorrect response. Despite being held back by a limited knowledge of the data schema, our results indicate that Text2Cohort always generates queries with correct structure and any queries with syntax or semantic errors can be corrected by an expert with minimal character-edits.

Recently, the paradigm of in-context learning has enabled zero-shot fine-tuning of LLMs using contextual information as a method for supervision \cite{petroni2020context,rau2023context,shi2023large,yang2022empirical,min2022rethinking}. For example, the open source package Llamaindex implements data connectors to pass various data sources as context to a GPT model \cite{liu2022llama}. For future work, we intend to explore these in-context learning techniques to address these limitations in Text2Cohort, while comparing with other state-of-the-art language models. Furthermore, we intend to release our dataset of natural language user inputs for others in the research community to also experiment with other techniques and languages models.

\section{Compliance with ethical standards} \label{sec:ethics}

This research study was conducted retrospectively using human subject data made available open access by the National Cancer Institute's Imaging Data Commons (\url{https://portal.imaging.datacommons.cancer.gov/explore}). Ethical approval was not required as confirmed by the license attached with the open access data.

\section{Acknowledgments}

No funding was received for conducting this study. The authors have no relevant financial or non-financial interests to disclose.

\bibliographystyle{IEEEbib}
\bibliography{refs}

\begin{thebibliography}{10}

\bibitem{fedorov2021nci}
Andrey Fedorov, William~JR Longabaugh, David Pot, David~A Clunie, Steve Pieper, Hugo~JWL Aerts, Andr{\'e} Homeyer, Rob Lewis, Afshin Akbarzadeh, Dennis Bontempi, et~al.,
\newblock ``Nci imaging data commons,''
\newblock {\em Cancer research}, vol. 81, no. 16, pp. 4188, 2021.

\bibitem{grossman2016case}
Robert~L Grossman, Allison Heath, Mark Murphy, Maria Patterson, and Walt Wells,
\newblock ``A case for data commons: toward data science as a service,''
\newblock {\em Computing in science \& engineering}, vol. 18, no. 5, pp. 10--20, 2016.

\bibitem{jiang2022big}
Peng Jiang, Sanju Sinha, Kenneth Aldape, Sridhar Hannenhalli, Cenk Sahinalp, and Eytan Ruppin,
\newblock ``Big data in basic and translational cancer research,''
\newblock {\em Nature Reviews Cancer}, vol. 22, no. 11, pp. 625--639, 2022.

\bibitem{khan2022transformers}
Salman Khan, Muzammal Naseer, Munawar Hayat, Syed~Waqas Zamir, Fahad~Shahbaz Khan, and Mubarak Shah,
\newblock ``Transformers in vision: A survey,''
\newblock {\em ACM computing surveys (CSUR)}, vol. 54, no. 10s, pp. 1--41, 2022.

\bibitem{brown2020language}
Tom Brown, Benjamin Mann, Nick Ryder, Melanie Subbiah, Jared~D Kaplan, Prafulla Dhariwal, Arvind Neelakantan, Pranav Shyam, Girish Sastry, Amanda Askell, et~al.,
\newblock ``Language models are few-shot learners,''
\newblock {\em Advances in neural information processing systems}, vol. 33, pp. 1877--1901, 2020.

\bibitem{liu2023pre}
Pengfei Liu, Weizhe Yuan, Jinlan Fu, Zhengbao Jiang, Hiroaki Hayashi, and Graham Neubig,
\newblock ``Pre-train, prompt, and predict: A systematic survey of prompting methods in natural language processing,''
\newblock {\em ACM Computing Surveys}, vol. 55, no. 9, pp. 1--35, 2023.

\bibitem{sun2023battle}
Shuo Sun, Yuchen Zhang, Jiahuan Yan, Yuze Gao, Donovan Ong, Bin Chen, and Jian Su,
\newblock ``Battle of the large language models: Dolly vs llama vs vicuna vs guanaco vs bard vs chatgpt--a text-to-sql parsing comparison,''
\newblock {\em arXiv preprint arXiv:2310.10190}, 2023.

\bibitem{rajkumar2022evaluating}
Nitarshan Rajkumar, Raymond Li, and Dzmitry Bahdanau,
\newblock ``Evaluating the text-to-sql capabilities of large language models,''
\newblock {\em arXiv preprint arXiv:2204.00498}, 2022.

\bibitem{kumar2022deep}
Ayush Kumar, Parth Nagarkar, Prabhav Nalhe, and Sanjeev Vijayakumar,
\newblock ``Deep learning driven natural languages text to sql query conversion: A survey,''
\newblock {\em arXiv preprint arXiv:2208.04415}, 2022.

\bibitem{yang2022empirical}
Zhengyuan Yang, Zhe Gan, Jianfeng Wang, Xiaowei Hu, Yumao Lu, Zicheng Liu, and Lijuan Wang,
\newblock ``An empirical study of gpt-3 for few-shot knowledge-based vqa,''
\newblock {\em Proceedings of the AAAI Conference on Artificial Intelligence}, vol. 36, no. 3, pp. 3081--3089, 2022.

\bibitem{dong2023c3}
Xuemei Dong, Chao Zhang, Yuhang Ge, Yuren Mao, Yunjun Gao, Jinshu Lin, Dongfang Lou, et~al.,
\newblock ``C3: Zero-shot text-to-sql with chatgpt,''
\newblock {\em arXiv preprint arXiv:2307.07306}, 2023.

\bibitem{choi2023chatgpt}
Jonathan~H Choi, Kristin~E Hickman, Amy Monahan, and Daniel Schwarcz,
\newblock ``Chatgpt goes to law school,''
\newblock {\em Available at SSRN}, 2023.

\bibitem{petroni2020context}
Fabio Petroni, Patrick Lewis, Aleksandra Piktus, Tim Rockt{\"a}schel, Yuxiang Wu, Alexander~H Miller, and Sebastian Riedel,
\newblock ``How context affects language models' factual predictions,''
\newblock {\em arXiv preprint arXiv:2005.04611}, 2020.

\bibitem{rau2023context}
Alexander Rau, Stephan Rau, Anna Fink, Hien Tran, Caroline Wilpert, Johanna Nattenmueller, Jakob Neubauer, Fabian Bamberg, Marco Reisert, and Maximilian~F Russe,
\newblock ``A context-based chatbot surpasses trained radiologists and generic chatgpt in following the acr appropriateness guidelines,''
\newblock {\em medRxiv}, pp. 2023--04, 2023.

\bibitem{shi2023large}
Freda Shi, Xinyun Chen, Kanishka Misra, Nathan Scales, David Dohan, Ed~Chi, Nathanael Sch{\"a}rli, and Denny Zhou,
\newblock ``Large language models can be easily distracted by irrelevant context,''
\newblock {\em arXiv preprint arXiv:2302.00093}, 2023.

\bibitem{min2022rethinking}
Sewon Min, Xinxi Lyu, Ari Holtzman, Mikel Artetxe, Mike Lewis, Hannaneh Hajishirzi, and Luke Zettlemoyer,
\newblock ``Rethinking the role of demonstrations: What makes in-context learning work?,''
\newblock {\em arXiv preprint arXiv:2202.12837}, 2022.

\bibitem{liu2022llama}
Jerry Liu,
\newblock ``{LlamaIndex},'' 11 2022.

\end{thebibliography}

\end{document}